\documentclass[sn-mathphys]{sn-jnl}

\jyear{2021}%

\usepackage{lineno,hyperref}
\modulolinenumbers[5]
\usepackage[T1]{fontenc}
\usepackage{graphics} 
\usepackage{epsfig} 
\usepackage{times} 
\usepackage{amsmath} 
\usepackage{amssymb}  
\usepackage{amsfonts}
\usepackage{bm}
\usepackage{algorithm, algpseudocode}
\usepackage{listings}
\usepackage{graphicx}
\usepackage{hyperref}
\usepackage{natbib}
\usepackage{wrapfig}
\usepackage{graphicx}
\DeclareUnicodeCharacter{2009}{\,}
\usepackage{multirow,tabularx}
\usepackage{subfigure}
\usepackage{booktabs}
\usepackage{xspace}

\usepackage{multirow}
\usepackage{amsmath,amssymb,amsfonts}
\usepackage{amsthm}
\usepackage{mathrsfs}
\usepackage[title]{appendix}
\usepackage{xcolor}
\usepackage{textcomp}
\usepackage{manyfoot}
\usepackage{booktabs}
\usepackage{algorithm}
\usepackage{algorithmicx}
\usepackage{algpseudocode}
\usepackage{listings}
\theoremstyle{thmstyleone}%
\theoremstyle{thmstyletwo}%

\theoremstyle{thmstylethree}%

\normalbaroutside

\begin{document}

\title[\tiny{Learning to Model Diverse Driving Behaviors in Highly Interactive Autonomous Driving Scenarios with Multi-Agent Reinforcement Learning}]{Learning to Model Diverse Driving Behaviors in Highly Interactive Autonomous Driving Scenarios with Multi-Agent Reinforcement Learning}


\author[1,2,4]{\fnm{Liu} \sur{Weiwei}} 

\author[2,3]{\fnm{Hu} \sur{Wenxuan}}

\author[4]{\fnm{Jing} \sur{Wei}}

\author[4]{\fnm{Lei} \sur{Lanxin}}


\author[4]{\fnm{Gao} \sur{Lingping}}

\author*[1]{\fnm{Liu} \sur{Yong}\email{yongliu@iipc.zju.edu.cn}}

\affil[1]{\orgdiv{the Advanced Perception on Robotics and Intelligent Learning Lab}, \orgname{College of Control Science and Enginneering, Zhejiang University}, \orgaddress{\city{Hangzhou}, \postcode{310027}, \country{China}}}

\affil[2]{\orgdiv{the Advanced Perception on Robotics and Intelligent Learning Lab}, \orgname{Huzhou Institute of Zhejiang University}, \orgaddress{\city{Huzhou}, \country{China}}}

\affil[3]{\orgdiv{College of Information Engineering, Huzhou University}, \orgaddress{\city{Huzhou}, \country{China}}}

\affil[4]{\orgdiv{Department of Autonomous Driving Lab}, \orgname{Alibaba DAMO Academy}, \orgaddress{\city{Hangzhou}, \country{China}}}

\abstract{Autonomous vehicles trained through Multi-Agent Reinforcement Learning (MARL) have shown impressive results in many driving scenarios. However, the performance of these trained policies can be impacted when faced with diverse driving styles and personalities, particularly in highly interactive situations. This is because conventional MARL algorithms usually operate under the assumption of fully cooperative behavior among all agents and focus on maximizing team rewards during training. To address this issue, we introduce the Personality Modeling Network (PeMN), which includes a cooperation value function and personality parameters to model the varied interactions in high-interactive scenarios. The PeMN also enables the training of a background traffic flow with diverse behaviors, thereby improving the performance and generalization of the ego vehicle. Our extensive experimental studies, which incorporate different personality parameters in high-interactive driving scenarios, demonstrate that the personality parameters effectively model diverse driving styles and that policies trained with PeMN demonstrate better generalization compared to traditional MARL methods.}


\maketitle

\section{Introduction}\label{sec1}

With the rapid development of Deep Reinforcement Learning (DRL) \cite{lillicrap2016continuous, peng2017multiagent, mnih2016asynchronous}, DRL has been widely used in the field of autonomous driving for decision-making and control\cite{shalev2016safe, bhalla2020deep, lei2021kb}. Moreover, due to the time-varying of vehicle interactions and the non-stationarity of the environment, it is challenging to design manual controllers in traffic scenarios. Therefore, people often use MARL algorithms \cite{zhou2022multi, dinneweth2022multi} to control vehicles to simulate the interaction behavior of autonomous vehicles. These trained policies \cite{he2023deep, dang2023event} are often used to directly control the ego vehicle, or serve as controllers for the background traffic flow.

\begin{figure}[htp]
\centering
\includegraphics[width=3.5in]{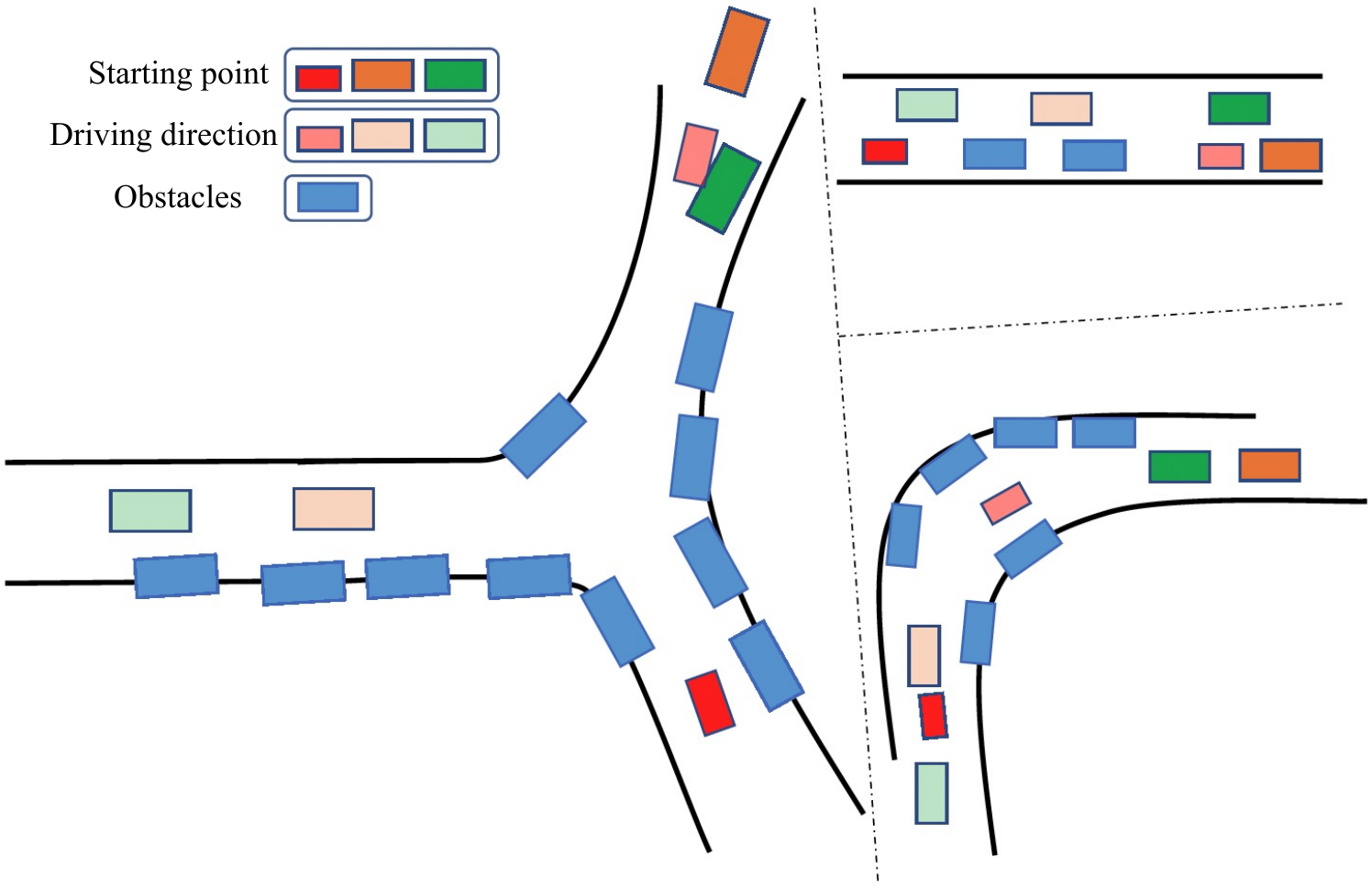}
\caption{Highly interactive driving scenarios: meeting with different road geometry.}
\label{asv_dynamic}
\end{figure}
Many previous works \cite{palanisamy2020multi, wachi2019failure} in Multi-Agent Reinforcement Learning (MARL) assume that all agents behave in a fully cooperative and rational manner in order to maximize the team's reward. Such algorithms tend to perform well in structured environments with strict rules \cite{lv2022safe}, however, the agents trained using these methods tend to display a solely cooperative behavior.

Therefore, there are two major challenges to implementing these methods in autonomous driving. Firstly, the existing reward assignment mechanisms \cite{neary2020reward} do not accurately model the cooperation levels of other drivers, causing the behavior of the trained agents to be monolithic and resulting in poor performance in high-conflict scenarios such as narrow residential roads (as illustrated in Figure 1) \cite{yaqoob2019autonomous}. Secondly, traffic participants often have different driving behaviors \cite{singh2021analyzing, bucsuhazy2020human}, while the absence of personality modeling in existing MARL methods can lead to over-fitting of trained policies and the inability to represent diverse driving scenarios in the real world. This results in a lack of training and testing for the behavior of the ego vehicle when using the trained agents as background traffic flow.

In order to enrich the driving style of the background traffic flow and train ego vehicles with the adaptive ability, while avoiding overfitting, some research efforts have been made \cite{ma2021reinforcement, wang2019achieving}. However, these methods are not suitable for modeling diverse behaviors in autonomous driving, because they either only be apply to simple discretized toy environments~\cite{wang2019achieving}, or fail to model the level of aggressiveness in a continuous manner~\cite{ma2021reinforcement}.

In this work, we propose the Personality Modeling Network (PeMN) to enhance the diversity of driving styles in both ego-vehicle and background traffic flows. PeMN incorporates personality parameters to separate the agent reward into individual and cooperative components, thus promoting independence in vehicle behavior. Additionally, PeMN models the state-action value function of other vehicles, allowing each vehicle to effectively monitor the behavior patterns of its peers. Our main contributions are summarized as follows:

\begin{itemize}

\item We propose PeMN to effectively map reward decoupling to behavior decoupling; the agent's value function is separated into a self-value function and a cooperation value function.

\item We use PeMN to model diverse driving styles through personality parameters, and the proposed PeMN is capable to create diverse interaction driving data. 

\item We conduct extensive experimental studies to explore and analyze modeling vehicle behaviors for autonomous driving in interactive scenarios. Additionally, we find that using background traffic with varying personalities to train the policy results in better performance.

\end{itemize}

\section{RELATED WORK}

Reinforcement learning (RL) involves iteratively interacting with simulated environments and learning through trial and error while guided by a reward function to train sequential decision-making models~\cite{icarte2018using}. This approach circumvents the need for costly expert sample collection or extensive parameter tuning of handcrafted controllers, making RL a popular choice in a variety of sequential decision-making applications, including games~\cite{mnih2015human, silver2016mastering}, autonomous driving~\cite{rhinehart2019precog, bouton2019reinforcement}, and recommendation systems~\cite{zhao2019deep}.

\subsection{RL and MARL Algorithms}

Initially, value-based~\cite{mnih2013playing} and policy-based~\cite{silver2014deterministic} reinforcement learning (RL) algorithms have been proposed. Subsequently, to address the challenges faced by policy gradient algorithms, such as the difficulty in determining step size, algorithms like Trust Region Policy Optimization(TRPO)~\cite{schulman2015trust} and Proximal Policy Optimization(PPO)~\cite{schulman2017proximal} are introduced.
Directly applying single-agent RL algorithms to multi-agent systems can be challenging due to non-stationarity. Independent Proximal Policy Optimization(IPPO)~\cite{de2020independent} has shown promise in some scenarios, but may be limited by learning instability theory and suboptimal observations.  
Multi-Agent Proximal Policy Optimization(MAPPO)~\cite{yu2021surprising} addresses multi-agent problems by training a centralized evaluation function that can assess each agent's performance from a global perspective, effectively mitigating non-stationarity. Multi-Agent Actor-Critic for Mixed Cooperative-Competitive Environments(MADDPG)~\cite{lowe2017multi} uses off-policy samples to improve training efficiency, but the performance is not as good as on-policy methods. 

For the credit assignment problem \cite{singh1996reinforcement}, \cite{wolpert2001optimal} of the multi-agent system, Value-DEcomposition Network(VDN) \cite{sunehag2018value} sums local value functions of agents to fit the joint value function, while Q-decomposition Multi-agent Independent eXtension(QMix) \cite{rashid2020monotonic} uses a mixing network to achieve a non-linear fit. counterfactual multi-agent(COMA) \cite{foerster2018counterfactual} uses counterfactual thinking to infer each agent's contribution to the team. 
However, these approaches 
aim to maximize team reward rather than diverse behaviors, while our work pursues diverse driving behaviors with different personalities, and improves the driving policy with diverse behaviors.

\subsection{Diverse Behaviour Modeling of Agents with RL}

Traditional RL algorithms have been widely used to train agents, but they often suffer from overfitting to fixed behavior patterns. This can lead to suboptimal performance when agents interact with actual vehicles, which exhibit varying levels of cooperation in real-world scenarios. To address this issue, researchers have proposed different approaches. For instance, a continuous hidden Markov model can be used to predict high-level motion intent and low-level interaction intent \cite{song2016intention}, thus mitigating overfitting. Alternatively, some researchers~\cite{kawamura2014action} have described willingness to cooperate as a degree of cooperation, using the distance between agents to determine the level of cooperation. Another approach, STGSage~\cite{ma2021reinforcement}, extracts two latent states of the agent: aggressive and conservative driving. However, these approaches are limited by a fixed cooperative level for the background traffic flow, which hinders the algorithm's ability to generalize to diverse scenarios. To overcome this limitation, researchers~\cite{wang2019achieving} have proposed the concept of Sequential Prisoner's Dilemmas (SPD)~\cite{hao2015introducing}, which trains two levels of cooperation policies and generates other degrees of cooperation by combining the two policies. However, SPD only works in the grid world. Other researchers~\cite{killing2021learning} have trained policies with different degrees of willingness to cooperate, but they rely on discretizing driving behavior into three fixed actions, which severely limits the algorithm's ability to generalize to different scenarios. In contrast, our proposed algorithm is completely end-to-end and avoids the underlying controller, thus reducing the need for parameter tuning. We also demonstrate that incorporating personality parameters into the reward function is essential for generating rich behaviors in vehicles.

\subsection{Application of RL in autonomous driving}
In the application of Reinforcement Learning (RL) in autonomous driving, \cite{sallab2016end} utilized the TORCS simulator to introduce a deep reinforcement learning system for predicting both discrete actions (DQN) and continuous actions (DDAC). In \cite{wang2018reinforcement}, the authors employed Q-learning to develop a policy for controlling autonomous vehicles to execute maneuvers such as maintaining a straight course, changing lanes (left/right), and adjusting speed.
In the work of \cite{ngai2011multiple}, the authors proposed learning multi-objective reinforcement learning strategies through Q-learning or Dual-Action Q-learning (DAQL). The latter makes individual action decisions based on whether other vehicles interact with a specific goal-oriented agent.
Moreover, multi-agent reinforcement learning can be applied to high-level decision-making and coordination among groups of autonomous vehicles. For instance, scenarios like overtaking on a highway \cite{yu2020distributed} or navigating through an intersection without signal control. Another application for multi-agent reinforcement learning methods is in the development of adversarial agents for testing autonomous driving strategies before deployment, as described in \cite{wachi2019failure}. These agents control other vehicles in a simulated environment to learn how to expose behavioral weaknesses in autonomous driving strategies through irregular or road-rule-breaking behavior. Finally, multi-agent reinforcement learning methods may play a crucial role in formulating safety strategies for autonomous driving, as highlighted by \cite{shalev2016safe}.
In general, owing to the rapid development of neural networks, deep reinforcement learning has evolved into a potent learning framework with the capability to acquire complex strategies in high-dimensional environments. This aligns naturally with the context of autonomous driving applications, and the amalgamation of these two elements has resulted in a dynamically emerging field.

\section{BACKGROUND}
\subsection{Reward Determines Behavior}
In a fully cooperative multi-agent system, the objective of the reinforcement learning algorithm is to maximize the team's long-term return~\cite{puterman1990markov, agarwal2020optimality}. This means that the optimal action for each agent is determined by the long-term team return. As a result, agents are willing to prioritize the team's interests over their individual goals, promoting teamwork and collaboration.
For agent $i$ in such a system, its long-term return is:
\begin{align}
\begin{split}
G_t=r_{t+1}+r_{t+2}+\dots = {{\mathop{ \sum }\limits_{k=0}^{ \infty }{ \lambda^{k}r_{t+k+1}}}},
\end{split}
\end{align}
where $r_{t+1}$ is the team reward at time $t+1$. $\lambda$ is the discount factor.
Estimating the long-term return of an agent in a state is challenging since the actual future return is unknown. To overcome this, the value function $V_i := V(s_1, s_2, \cdots, s_n; r_i)$ is used. This function provides an estimate of the long-term return for the agent in state $S=\{s_1, s_2, \cdots, s_n\}$. The value function is dependent on the reward function $r_i$, which influences the behavior of agent $i$. In fully cooperative tasks, the team reward function is used as $r_i$. However, the optimal team reward for each task is unique, leading to a fixed behavior pattern for agents. To introduce diversity in agent behavior patterns, adjustment on the reward function $r_i$ is needed to diversify the agent's policy.

As depicted in Figure 2, the process of personality modeling comprises two main components. Firstly, the action-value function of the intelligent agent is expanded using a Taylor series into self-action value function and cooperative action value function. Subsequently, the reward for the intelligent agent is decomposed into self-reward and cooperative reward. This decomposition is employed to provide supervision signals for the self-action value function and cooperative action value function, respectively. Detailed descriptions of these two components are presented in Sections 3.2 and 3.3, respectively.

\section{METHODS}
\subsection{Decompose action-value function}

In order to realize the decoupling of the agent's actions from the team and its interests, we divide the reward $r_i$ of agent $i$ into two parts. The first part is the cooperative reward $r^{c}_{i}$, which contributes to the team's overall benefits. The second part is the self-reward $r^s_i$, which reflects the agent's inherent greed. The proportion of the two rewards can be adjusted to modify the agent's contribution to the team or itself. The agent's self-reward coefficient $\alpha$ is a personality parameter that balances the relationship between the two rewards. The agent takes action based on its current state $s_i$. The state value function $V_i$ is a function of the state $s_i$ and the rewards $r^s_i$ and $r^{c}_i$. The state-action value function $Q_i$ is a function of both the state $s_i$, the action taken by the agent $a_i$, and the rewards $r^s_i$ and $r^{c}_i$.

The personality of the agent determines the extent to which its behavior benefits the team and itself, which allows for the decoupling of the agent's behavior. Therefore, the objective function that is maximized can be expressed as follows:
\begin{align}
\begin{split}
J (r_1,r_2, \cdots ,r_n):=\sum_{i=1}^{n}{V_i (s_i; r_i^s, r_i^c)},\\
s.t.{ r_i^c  \le r_T}.
\end{split}
\end{align}
The equation above shows that the single-step reward $r_i$ of agent $i$ can be decomposed into its self-reward $r^s_i$ and cooperative reward $r^{c}_i$. This decomposition includes a constraint that limits the agent's contribution to the team to an upper bound of the team reward $r_T$. Equation 2 shows that the behavior of the agent is influenced by both the self-reward $r^s_i$ and the cooperative reward $r^c_i$. Specifically, when we increase $r^c_i$ and decrease $r^s_i$, the agent's behavior becomes more cooperative, and vice versa.
\begin{figure}[!t]
\centering
\includegraphics[width=3.5in]{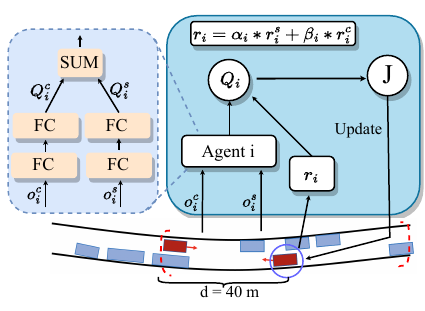}
\caption{The algorithm framework of PeMN.}
\label{asv_dynamic}
\end{figure}
While adjusting the personality parameters can produce diverse behavior patterns, we can better model the cooperation rewards offered by the team by utilizing the Taylor expansion at the ground-zero reward $r_i^s=r_i(s_i)$. This corresponds to the team reward when only agent $i$ is present in the environment  
and $r_i=r_{i}^{s}+r_{i}^{c}$:
\begin{align}
\begin{split}
&Q_i(s_i,a_i; r_i) = Q_i(s_i,a_i; r_i^s)+\\
& \nabla_r Q_i(s_i, a_i; r_i^s)(r_i - r_i^s) + \rm \mathcal{O}( \left\Vert r_i - r_i^s\right\Vert ^2 ).
\end{split}
\end{align}

The first term of the above equation is the policy adopted by the agent $i$ for its own reward $r_i^s$. we name it $Q^s_i= Q_i(s_i,a_i; r_i^s)$, and the second part is what the agent does for team cooperation. we name it $Q^c_i(s_i, a_i, r^{c}_{i})=\nabla_r Q_i(s_i, a_i; r_i^s)(r_i - r_i^s) + \rm \mathcal{O}( \left\Vert r_i -r_i^s \right\Vert ^2 )$. $Q^s_i$ and $Q^c_i$ are fitted by the neural network and we have:
\begin{align}
\begin{split}
Q_i(o_i, a_i) = Q^s_i(o^s_i,a_i; r_i^s) + Q^c_i(o^{c}_i,a_i;r_i^c),
\end{split}
\end{align}
Since the team reward depends on the state of other agents, $o^{s}_i$ denotes the observation of the agent i. $o^{c}_i$ is used here to represent the joint observation of the other agents in the team. 
The architecture of the PeMN is shown in Figure 2.

\subsection{Personality Parameterized Reward Function}

In the previous section, the action value function is decomposed into self-value function and cooperation value function, which are used to evaluate the two policies of the agent, and self-reward and cooperation reward are designed.
This section aims to apply personality modeling to the field of autonomous driving, reduce agent exploration time and improve algorithm convergence,
we designed dense and sparse rewards based on expert knowledge:
\begin{align}
\begin{split}
r_d\! =\! k_1(d_{t-1}^{pos}-d_t^{pos})\! +\! k_2 v_t / v_{max}
\end{split}
\end{align}
\begin{align}
\begin{split}
r\mathop{{}}\nolimits_{{p}}={ \left\{ \begin{array}{*{20}{l}}{20, \qquad  \qquad  \qquad if \, reach \, goal}
\\{-30,  \qquad \qquad  \, \quad if \, collision}
\\{-30,\qquad \qquad  \, \quad if \, out \, of \, raod}
\\{0,\qquad \qquad \qquad  \,\, if \, timeout}\end{array}\right. }
\end{split}
\end{align}

where $r_d$ is a dense reward to avoid the reward sparsity problem in reinforcement learning. $d_t^{pos}$ is the position on the vehicle lane at time $t$,  and $v_t$ is the vehicle speed and $v_{max}$ is the maximum speed limit in the lane. $k_1$ and $k_2$ are reward control coefficients. $r_p$ is the sparse reward, the purpose of which is to inform the agent that the current state is good or bad at the end of the episode. In this way, the self-reward of agent $i$ is designed as:
\begin{align}
\begin{split}
r_i^s = r_d + r_p.
\end{split}
\end{align}

Second, the personality parameter $\alpha$ is introduced into the reward function for modeling the vehicle's personality. Therefore, the reward function of the designed agent $i$ is: 
\begin{align}
\begin{split}
{ \left\{ \begin{array}{*{20}{l}}{r_i = \alpha_i * r^{s}_i + \beta_i * r^{c}_{i},
} \;\;\;if \; d  \le  40 \; m\\{r_i=r^{s}_i.} \qquad\qquad\qquad\;\;\;if \; d > 40\; m \end{array}\right. }
\end{split}
\end{align}
To ensure that the vehicle retains its ability to target and avoid obstacles, we excluded the reach goal reward and collision penalty from the reward distribution formula described above. The team reward for all vehicles within range, except for the vehicle itself (i.e., $r^{c}_i = r^s_j + \cdots + r^s_k, i \neq k$), is used to calculate the vehicle's reward. Additionally, we use the parameters $\alpha$ $(\alpha \geq 0)$ and $\beta$ $(\beta \geq 0)$ to control the vehicle's self-reward and cooperation reward, respectively, and to define its personality. Typically, we set $\alpha + \beta = 1$ to ensure that the vehicle is indifferent to external rewards when its personality is completely selfish, and vice versa. Finally, Equation 8 shows that as the personality parameter decreases, the degree of cooperation among vehicles improves.

\subsection{Training}
As distributed deployment is required, we follow a centralized training and decentralized execution (CTDE) framework. This means that during training, we can utilize information from other vehicles, while during execution, each vehicle only relies on its own state. To accomplish this, we utilize the MAPPO algorithm for policy iteration, with the objective function defined as follows:

\begin{align}
\begin{split}
J_{\theta _k}( \theta ) \approx \varSigma _{( s_t, a_t )}\min ( \frac{p_{\theta}( a_t, s_t )}{p_{\theta _k}( a_t, s_t )}A_{\theta _k}( a_t\mid s_t ) , \\
cilp( \frac{p_{\theta}( a_t, s_t )}{p_{\theta _k}( a_t, s_t )}, 1-\varepsilon , 1+\varepsilon ) A_{\theta _k}( a_t\mid s_t ) ) 
\end{split}
\end{align}
here:
\begin{align}
\begin{split}
& A{ \left( {s\mathop{{}}\nolimits_{{t}},a\mathop{{}}\nolimits_{{t}}} \right) }=Q{ \left( {s\mathop{{}}\nolimits_{{t}},a\mathop{{}}\nolimits_{{t}}} \right) }-V{ \left( {s\mathop{{}}\nolimits_{{t}}} \right) }\\=
& E\mathop{{}}\nolimits_{{s\mathop{{}}\nolimits_{{t+1}} \backsim p{ \left( {s\mathop{{}}\nolimits_{{t+1}} \mid s\mathop{{}}\nolimits_{{t}}, a\mathop{{}}\nolimits_{{t}}} \right) }}}{ \left[ {r{ \left( {s\mathop{{}}\nolimits_{{t}}} \right) }+ \gamma V\mathop{{}}\nolimits^{{ \pi }}{ \left( {s\mathop{{}}\nolimits_{{t+1}}} \right) }-V\mathop{{}}\nolimits^{ \pi }{ \left( {s\mathop{{}}\nolimits_{{t}}} \right) }} \right] }
\end{split}
\end{align}

\begin{align}
\begin{split}
Q(s_t, a_t) = Q^s(o^s_t,a_t; r^s_t) + Q^c(o^{c}_t,a_t;r^c_t),
\end{split}
\end{align}

Here, $A(s_t, a_t)$ is the advantage function, which represents the difference in magnitude between the agent's action-value function and the state-value function. It is used to assess whether the actions taken by the agent in this state are beneficial. $Q(s_t, a_t)$ is the agent's action-value function, and $V(s_t)$ is the agent's state-value function. $\frac{p_{\theta}( a_t, s_t )}{p_{\theta _k}( a_t, s_t )}$ is the ratio between the new policy and the old policy, indicating the change in probability of the new policy along the sampled trajectory compared to the old policy. $\epsilon$ is a clipping parameter used to ensure that the magnitude of policy updates does not exceed a threshold. $\gamma$ is the discount factor.The pseudo code of the algorithm is as follows:

\begin{algorithm}
\caption{PeMN}
\begin{algorithmic}
\For{$iteration=1$, $2$, $\ldots$} \label{alg:ppo_iteration}
\For{$actor=1$, $2$, $\ldots$, $N$} \label{alg:ppo_actor}
    \State Run policy $\pi_{\theta_{old}}$ in environment for $T$ timesteps
    \State Obtain observation $o_{t} = o_{t}^s + o_{t}^c$ from the environment
    \State obtain the reward $r_{t}$ according to equation 8
\State Calculate advantage estimates $A_{1}, \ldots, A_{T}$ using equation 10 and 11
\EndFor

\State Optimize surrogate $L$ wrt $\theta$, with $K$ epochs and minibatch size $M \leqslant NT$
\State $\theta_{old} \leftarrow \theta$
\EndFor

\end{algorithmic}
\end{algorithm}

Once we have finished modeling the vehicles with multiple personalities, we obtain models with rich and diverse behavioral patterns. We then use the models to create the traffic background flow. This enables us to train an ego vehicle with adaptive cooperative abilities. Although we are able to discern the individual personalities of the other agents, but instead of feeding this personality as an observation into the network, we continue to rely on the centralized training with decentralized execution approach to train the ego vehicle. This ensures that the ego car can operate effectively in complex traffic scenarios, while adapting to the diverse and unpredictable behaviors of other agents.

\section{Results and Discussions}
\begin{figure*}[b]
\centering
\includegraphics[width=4.5in]{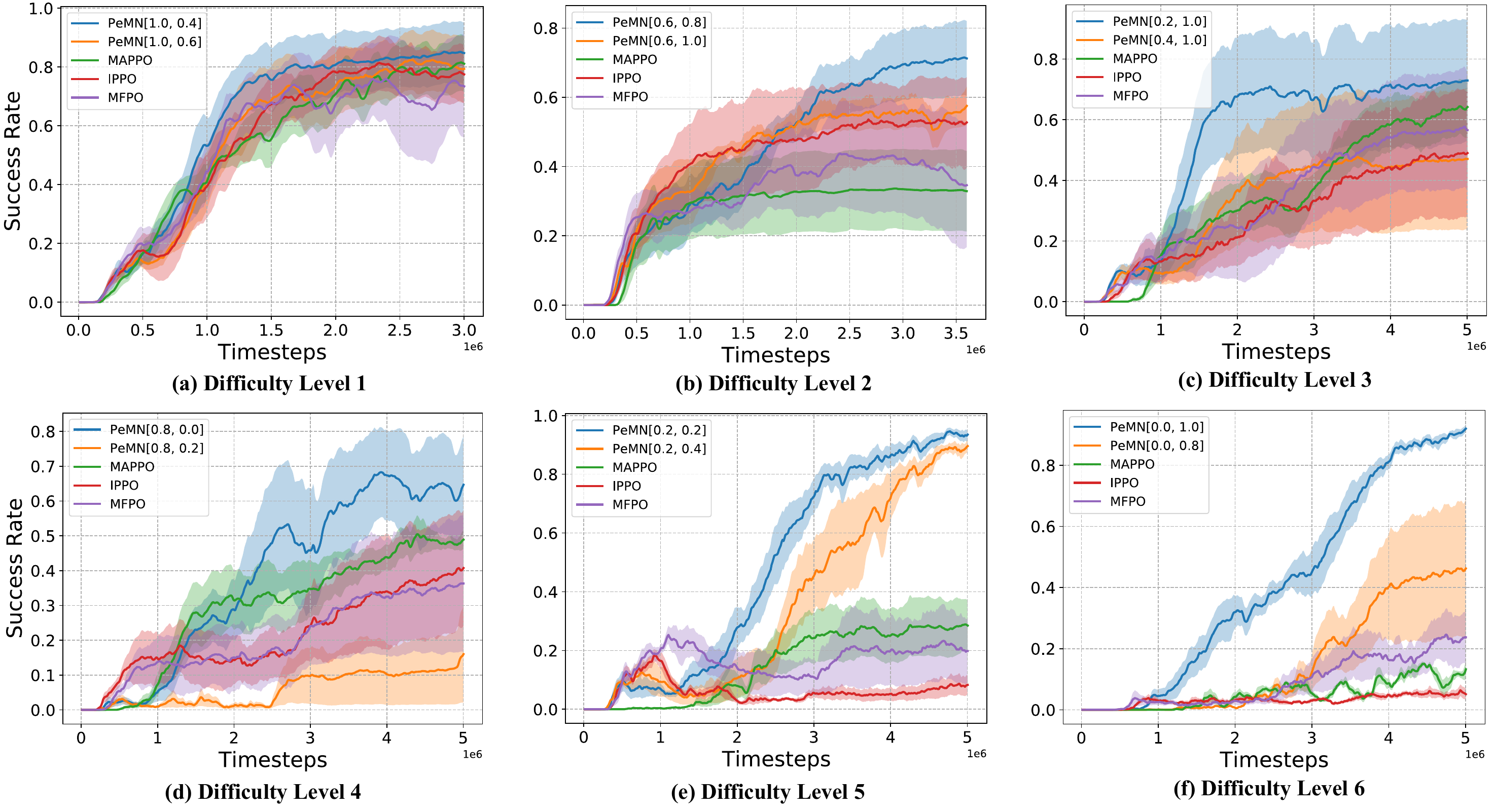}
\caption{Performance of PeMN and baseline algorithms in different scenarios. In the figure, PeMN[1.0, 0.4] represents the personality parameters $\alpha_l, \alpha_r$ of the left and right cars are 1.0 and 0.4, respectively. [1, 0.4] is a personality pair.
}
\label{asv_dynamic}
\end{figure*}

\subsection{Experiment Setup}
We utilize RLlib~\cite{liang2018rllib} as our training framework and compare our approach to existing methods: centralized training with decentralized execution (MAPPO, MFPO) and decentralized training with decentralized execution (IPPO). MAPPO is a PPO algorithm variant with a centralized value function for MARL, MFPO~\cite{yang2018mean} is a Mean Field Policy Optimization algorithm based on mean field theory, and IPPO employs PPO as individual learners. Hyperparameters used include a learning rate of 1e-4, a discount factor of 0.9, a sample batch size of 10240, and 20 processes for sample collection. We train our model using 8 Nvidia RTX 3090 GPUs and 44 CPUs. The performance and effectiveness of our method are evaluated on the narrow road meeting scenario, which is challenging due to limited lane width, obstacles, highly interaction. We use the open-source simulator MetaDrive \cite{li2022metadrive} to create six difficulty levels based on obstacle density and road geometry. This is a study that supports the generation of infinite scenarios with various road maps and traffic settings for generalizable RL. The emulator features accurate physics simulation and multiple sensory inputs including lidar, RGB images, and more. Here, information from radar perception and the vehicle's own state (For details, please refer to the open source library MetaDrive.) are used as agent observations. The PeMN output continuous actions are the throttle (or brake) and steering of the vehicle.

\subsection{Results Comparison}

Personality parameters are assigned to the left and right cars in Figure 4, forming a personality pair.
PeMN demonstrates higher success rates, especially within well-matched personality pairs, as shown in Figure 3, even amid escalating scenario complexity. This effectiveness arises from PeMN's ability to collaborate and comprehend opponent behavior through the value function. The suitable personality facilitates reward function decomposition, guiding the vehicle safely through the avoidance area while considering other vehicles' behavior. Conversely, an unsuitable personality may impede vehicle cooperation. Maximum success rate personality pairs vary across scenarios due to different terrains.

\begin{figure}[h]
\centering
\includegraphics[width=4.5in]{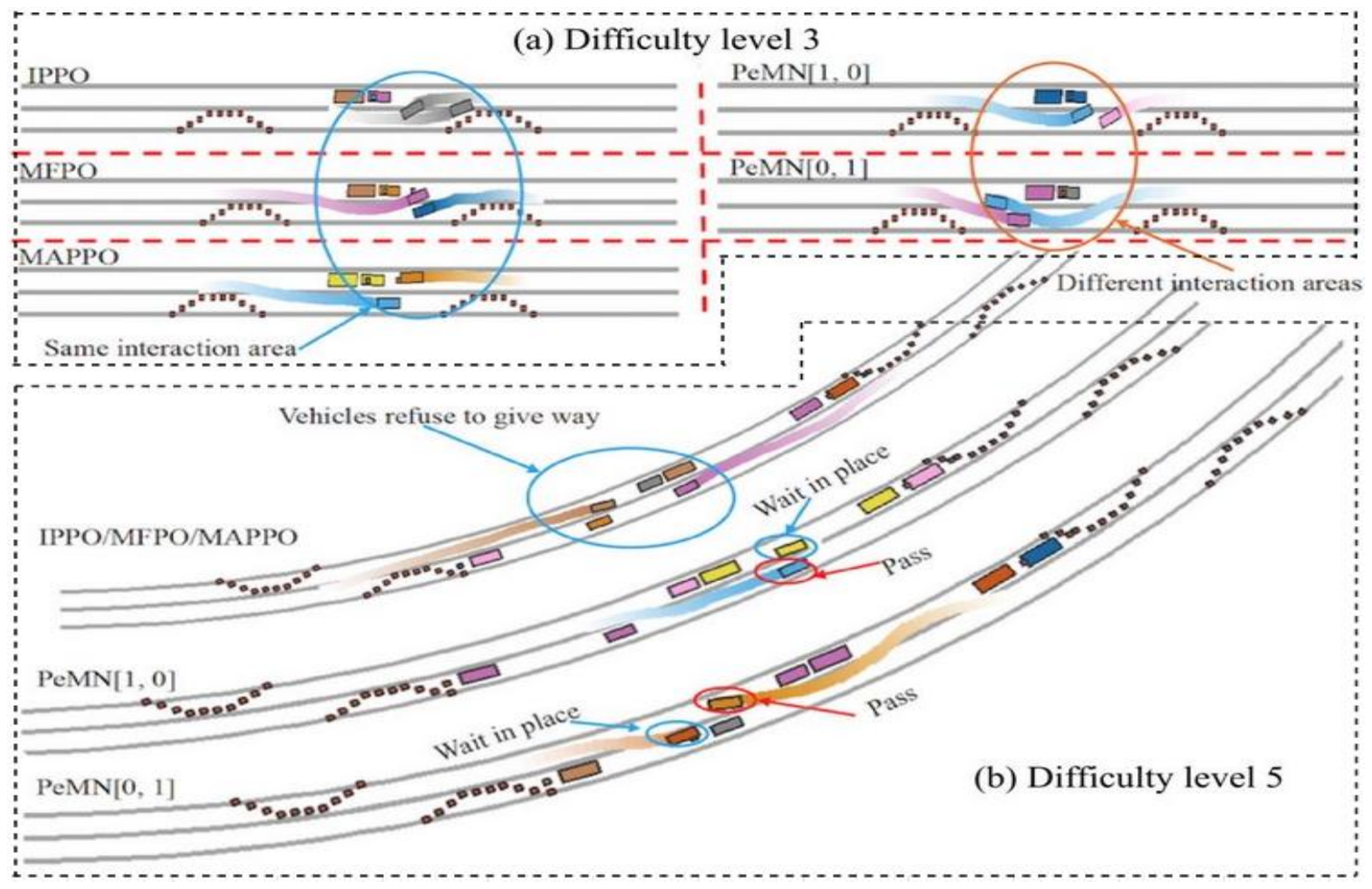}
\caption{
The simulated vehicle trajectory is visualized to demonstrate the disparities in driving behavior between PeMN and baseline algorithms. The color from dark to light indicates the position of the vehicle in the past 25 steps.
}
\label{asv_dynamic}
\end{figure}

In Figure 4a, the baseline algorithm displays a uniformly defined avoidance area due to its prioritization of maximum returns. The Nash equilibrium point is positioned to the right of the straight road, resulting in a fixed pattern of aggressive driving on the left and cautious driving on the right, despite the vehicle's capability to safely pass through. Unfortunately, collisions occur when the left car encounters the aggressively driving right car, leading to mission failure.
In Figure 4b, scenarios reveal challenges with the baseline algorithm's normal reward function guiding the vehicle toward exploring the avoidance area, resulting in an unwillingness to avoid collisions or road blockages. In contrast, PeMN presents a distinct approach. Figure 4 illustrates a scenario where left and right vehicles embody extreme personalities – conservative and aggressive, respectively. The outcomes show interactions in both left and right avoidance areas, fostering diverse driving styles. The key takeaway is that personality significantly influences the creation of varied driving styles and the diversification of interaction data.

\begin{figure*}[h]
\centering
\includegraphics[width=\textwidth]{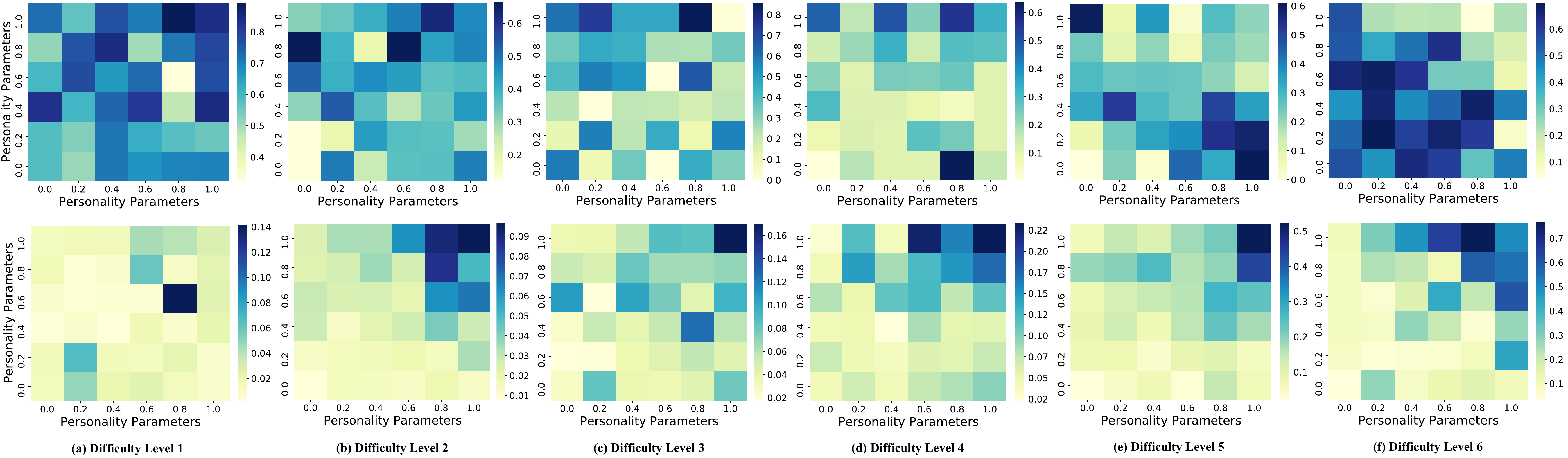}
\caption{The difference between the success rate and the collision rate with different personality pairs. Darker color represents higher success rate or collision rate.}
\label{asv_dynamic}
\end{figure*}

\subsection{Diverse Behaviours and Diverse Interaction Data with Personality Differences}

The previous section explored how different personality pairs influence vehicle interactions in various regions. This section analyzes additional indicators of personality differences. Figure 5's first row suggests that while the success rate of a given personality pair varies across scenarios, there are no consistent patterns. This emphasizes the strong correlation between terrain scenarios (e.g. road geometry/topology/elements) and personality performance. It disproves the stereotype that conservative driving styles work in any situation, as in the RL/MARL policies for autonomous driving. Although optimal personality pairs differ across scenarios, success rates of different personality pairs in the same scenario follow regular patterns. The first row of Figure 5 demonstrates that the success rate generally increases as the personality pair gets closer to the optimal personality pair for that scenario.
\begin{figure}[h]
\centering
\includegraphics[width=3in]{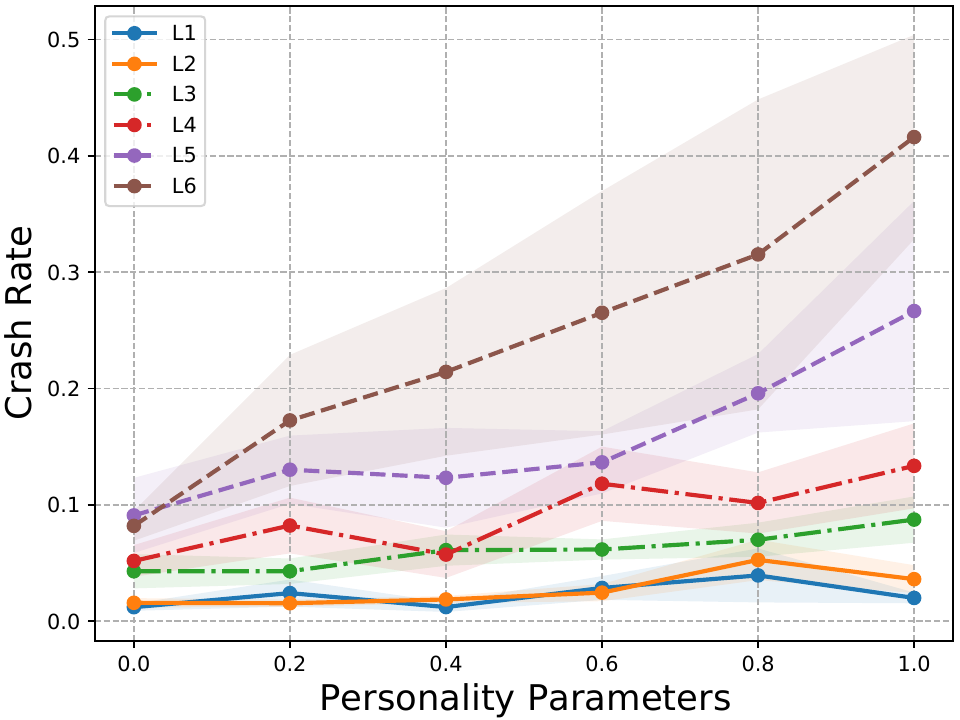}
\caption{Analysis of crash rate and personality parameters. $L_i$ represents difficulty level $i$.}
\label{asv_dynamic}
\end{figure}
\begin{figure}[h]
\centering
\includegraphics[width=4in]{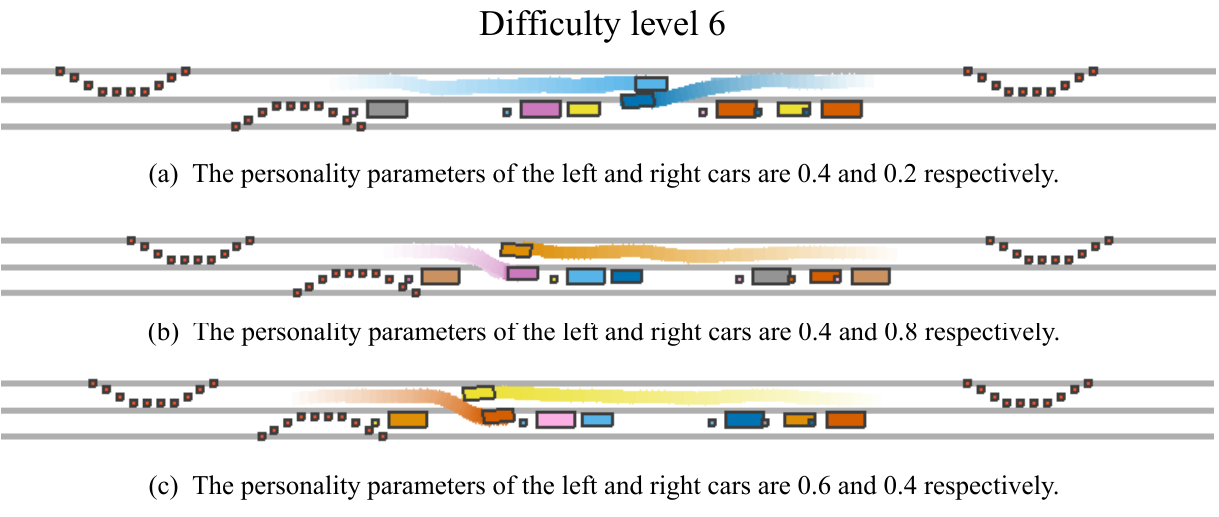}
\caption{PeMN modeling driving behaviors in interactive scenarios. The interactions are affected by road elements and geometry, and diverse interaction data are created through behavioral modeling.}
\label{asv_dynamic}
\end{figure}

The impact of personality on collision rates in diverse scenarios is noteworthy. Figure 6 illustrates that elevating the personality parameter correlates with heightened aggressive driving tendencies and reduced attention to other vehicles, consequently escalating collision rates. Additional experiments involved keeping the left car's personality constant while varying the right car's personality during training. Results indicated a consistent rise in collision rates as the personality parameter of the left car increased, irrespective of the scenario. Notably, this effect intensified with a greater number of obstacles. In summary, irrespective of terrain scenarios, higher levels of vehicle aggressiveness in terms of personality lead to an increase in collision rates.

Figure 7a-b illustrates varying performance levels among different vehicle personalities. In 7a, the left car showcases a more aggressive driving style with a larger personality parameter, contrasting with the right car's more conservative behavior. However, in 7b, the left car adopts a more conservative approach when confronted with an even more aggressive personality on the right. Importantly, such situations are not always straightforward. In 7c, despite possessing a more aggressive personality than the right car, the left car embraces a conservative driving style. This is attributed to a relatively small personality gap, where the weaker personality parameter prompts the left car to prioritize gaining a higher reward by increasing speed. Additionally, the road topology and layout favor the left car to decelerate and avoid collisions. Moreover, the reward for braking the right car is smaller compared to braking the left car. Consequently, we can conclude that the relative aggressiveness of different personality parameters is influenced by road topology and geometry.

\subsection{Adaptive Cooperation with Traffic Flow Modeled with Different Personalities}

Results in Table 1 and Figure 8 demonstrate that the behavioral modeling with the personality parameters improve the training. In this experiment, we utilized a diverse set of background traffic flows with different driving styles to train our adaptive cooperation algorithm. Specifically, we employed various "personalities" to represent different driving styles during training. Additionally, we conducted experiments using three difficulty levels of straight roads.
In our first set of experiments, we trained the background traffic flow using 6 sets of personality pairs, each representing a unique driving style. We then used this trained background traffic flow as the interactive object for training our proposed algorithm. To evaluate the efficacy of our approach, we conducted experiments using a MARL baseline algorithm to control the vehicles during training. Our results, presented in Table 1, demonstrate that the MARL baseline algorithm often falls into a fixed driving style. 
Overall, our experiments underscore the importance of incorporating diverse driving styles into the training of cooperative driving algorithms. By doing so, we can improve their adaptability and robustness in real-world scenarios.
The three dimensions of the comparison in Figure 8 are: \textit{success rate} is the vehicle success rate, \textit{safety} is the vehicle non-collision rate, \textit{efficiency} $ >= 0$ indicates the difference between successes and failures in a unit of time $\left( N\mathop{{}}\nolimits_{{success}}-N\mathop{{}}\nolimits_{{failure}} \left) /T\right. \right.$. 
The adaptive cooperative algorithm is trained using PeMN without reward decomposition.

\begin{table*}[h]
\caption{Comparison of success rate among the baseline methods and our method. Background traffic flow with different driving styles are used to evaluate the policies, each row represents a traffic flow with a specific personality or was tested using existing RL/MARL methods.}
\centering
\resizebox{\textwidth}{20mm}{
\begin{tabular}{ccccccccccccc}
\hline
 \multicolumn{1}{l}{}       & \multicolumn{4}{c}{Difficulty Level 1}                                                                    & \multicolumn{4}{c}{Difficulty Level 3}                                                                  & \multicolumn{4}{c}{Difficulty Level 6}                                               \\ \hline
                           & IPPO             & MAPPO            & MFPO             & Ours                               & IPPO             & MAPPO            & MFPO             & Ours                               & IPPO             & MAPPO            & MFPO             & Ours          \\ \hline
\multicolumn{1}{c|}{0.0}   & 0.61             & \textbf{0.99}    & 0.01             & \multicolumn{1}{c|}{0.85}          & 0.01             & \textbf{0.51}    & 0.0              & \multicolumn{1}{c|}{0.26}          & 0.01             & 0.06             & 0.39             & \textbf{0.48} \\
\multicolumn{1}{c|}{0.2}   & 0.43             & \textbf{0.95}    & 0.0              & \multicolumn{1}{c|}{\textbf{0.95}} & 0.29             & 0.59             & 0.04             & \multicolumn{1}{c|}{\textbf{0.6}}  & 0.32             & 0.13             & 0.53             & \textbf{0.69} \\
\multicolumn{1}{c|}{0.4}   & 0.32             & 0.57             & 0.48             & \multicolumn{1}{c|}{\textbf{0.80}} & 0.36             & 0.59             & 0.05             & \multicolumn{1}{c|}{\textbf{0.85}} & 0.27             & 0.16             & 0.52             & \textbf{0.70} \\
\multicolumn{1}{c|}{0.6}   & 0.36             & 0.52             & 0.49             & \multicolumn{1}{c|}{\textbf{0.93}} & 0.27             & 0.57             & 0.05             & \multicolumn{1}{c|}{\textbf{0.76}} & 0.02             & 0.14             & \textbf{0.67}    & 0.48          \\
\multicolumn{1}{c|}{0.8}   & 0.02             & 0.33             & \textbf{0.92}    & \multicolumn{1}{c|}{\textbf{0.92}} & 0.29             & \textbf{0.70}    & 0.36             & \multicolumn{1}{c|}{0.65}          & 0.10             & 0.28             & \textbf{0.57}    & 0.35          \\
\multicolumn{1}{c|}{1.0}   & \textbf{0.98}    & 0.89             & 0.05             & \multicolumn{1}{c|}{0.68}          & 0.38             & 0.15             & 0.21             & \multicolumn{1}{c|}{\textbf{0.60}} & 0.03             & 0.03             & 0.17             & \textbf{0.46} \\
\multicolumn{1}{c|}{IPPO}  & \textbackslash{} & \textbackslash{} & \textbackslash{} & \multicolumn{1}{c|}{\textbf{0.58}} & \textbackslash{} & \textbackslash{} & \textbackslash{} & \multicolumn{1}{c|}{\textbf{0.70}} & \textbackslash{} & \textbackslash{} & \textbackslash{} & \textbf{0.49} \\
\multicolumn{1}{c|}{MAPPO} & \textbackslash{} & \textbackslash{} & \textbackslash{} & \multicolumn{1}{c|}{\textbf{0.83}} & \textbackslash{} & \textbackslash{} & \textbackslash{} & \multicolumn{1}{c|}{\textbf{0.42}} & \textbackslash{} & \textbackslash{} & \textbackslash{} & \textbf{0.43} \\
\multicolumn{1}{c|}{MFPO}  & \textbackslash{} & \textbackslash{} & \textbackslash{} & \multicolumn{1}{c|}{\textbf{0.97}} & \textbackslash{} & \textbackslash{} & \textbackslash{} & \multicolumn{1}{c|}{\textbf{0.78}} & \textbackslash{} & \textbackslash{} & \textbackslash{} & \textbf{0.44} \\
\hline
\end{tabular}}
\end{table*}

\begin{figure*}[t]
\centering
\includegraphics[width=4in]{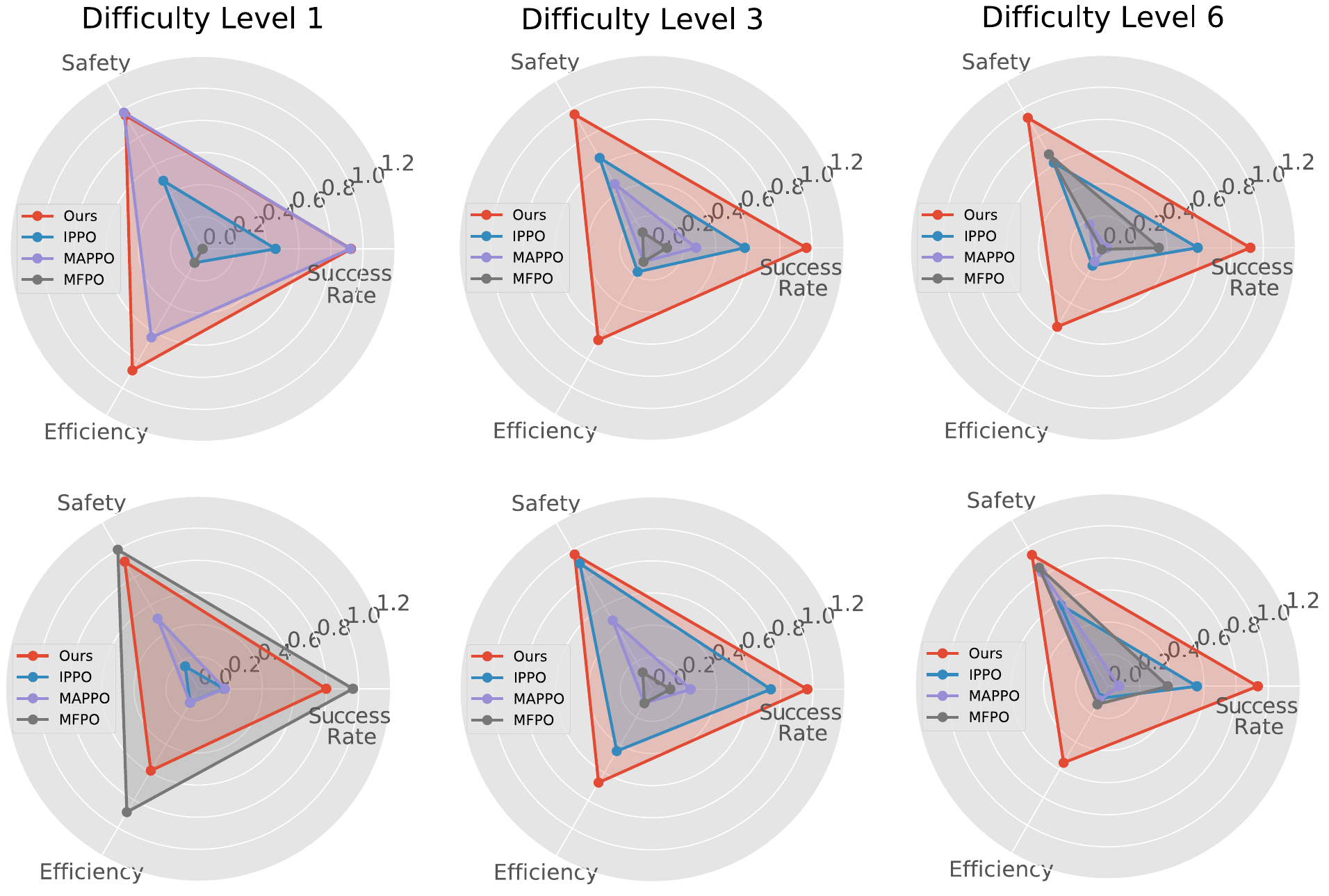}
\caption{The interaction performance of the trained algorithms (\textit{Ours}, the baseline) and the background traffic flow with two different driving styles in three dimensions. Note that the personality pairs for the initial training of the background traffic flow vehicles in the three figures (above) are [0.2, 0.2], and the three figures (below) are [0.4, 0.2].}
\label{asv_dynamic}
\end{figure*}

We employed various personalities and three baseline algorithms to control background traffic flow. In Table 1, the baseline algorithm excels against opponents with suitable personalities but significantly falters against uncooperative ones. Its lack of robustness in facing opponents with diverse personalities suggests challenges in generalization for existing MARL algorithms, despite strong training performance. In contrast, our algorithm maintains a consistently high success rate, demonstrating robustness against opponents with different driving styles and surpassing most other experiments. Importantly, for a fair comparison, we refrained from training our algorithm with the baseline algorithm; instead, we assessed its performance by testing it against the baseline algorithm. Even under these conditions, our algorithm consistently succeeded against opponents with different personalities. This underscores the efficacy of using background traffic flow with diverse driving styles to substantially enhance the generalization ability of ego vehicle control algorithms.

In addition, we evaluated the adaptive cooperation ability of our method in various scenarios, as presented in Figure 8. The results reveal excellent performance of our approach in three distinct difficulty levels and under two different driving styles of background traffic flow. Across the experiments, the adaptive cooperation algorithm consistently outperformed all other methods, achieving first place in all three dimensions. These results highlight the ability of our algorithm to adapt and generalize to diverse traffic scenarios, while simultaneously improving safety and efficiency. Notably, our approach also effectively handles background traffic flow with multiple driving styles, further demonstrating its versatility and robustness.

\section{Conclusions}
In this paper, we introduce the PeMN, a model for characterizing agent personalities. Additionally, we utilize the intricate behavior patterns of these vehicles as the underlying traffic flow to enhance the training of an ego vehicle with adaptive cooperative capabilities. The study of behavioral modeling of traffic participants, especially in highly interactive scenarios, remains a challenging and open research topic. Our work aims to tackle this challenge by exploring behavior modeling for diverse driving data generation and studying behaviors in highly interactive scenarios. We also demonstrate that our modeling method achieves better performance and generalization in highly interactive scenarios.

In the future, we will further investigate the connections between personality modeling networks and the well-known credit assignment problem in multi-agent reinforcement learning algorithms. This exploration aims to enhance the collaborative capabilities among intelligent agents.

\section{Acknowledgments}
This work was supported by the National Defense Basic
Research Program (JCKY2021204B051).

\bibliography{sn}

\end{document}